\documentclass[journal]{IEEEtran}
\usepackage{color, xcolor}         

\usepackage[namelimits]{amsmath} 
\usepackage{amssymb}             
\usepackage{amsfonts}            
\usepackage{mathrsfs}            
\usepackage{booktabs}
\usepackage{multirow}
\usepackage{stfloats}
\usepackage{bbding}
\usepackage{amssymb}  
\usepackage{float}
\usepackage[pdftex]{graphicx}
\usepackage{bm}
\usepackage{soul}
\definecolor{myblue}{RGB}{47,84,150}

\usepackage{cite}

\ifCLASSINFOpdf
\else
\fi
\begin{document}
%
\title{LASOR: Learning Accurate 3D Human Pose and Shape Via Synthetic Occlusion-Aware Data and Neural Mesh Rendering}
%
%
%
\author{\IEEEauthorblockN{Kaibing~Yang\IEEEauthorrefmark{1},
Renshu~Gu\IEEEauthorrefmark{1}, Maoyu~Wang, Masahiro~Toyoura,
Gang~Xu}\\
\IEEEauthorblockA{\IEEEauthorrefmark{1}Equal Contribution}
\thanks{Manuscript received July 7, 2021; revised December 10, 2021 and January 23, 2022. This work was supported by the National Key R\&D Program of China under Grant No.2020YFB1709402, the NSFC-Zhejiang Joint Fund for the Integration of Industrialization and Informatization (Grant No. U1909210), the Zhejiang Provincial Science and Technology Program in China under Grant No. 2021C01108, Zhejiang Lab Tianshu Open Source AI Platform, the Zhejiang Provincial Science and Technology Program in China under Grant No. LQ22F020026, and Fundamental Research Funds for the Provincial Universities of Zhejiang (Grant No. GK219909299001-028). (Corresponding authors: Gang Xu; Masahiro Toyoura.)}
\thanks{Kaibing Yang, Renshu Gu, Maoyu Wang, and Gang Xu are with the School of Computer Science and Technology, Hangzhou Dianzi University, Hangzhou 310018, China (e-mail: 191050068@hdu.edu.cn; renshugu@hdu.edu.cn; 202320053@hdu.edu.cn; gxu@hdu.edu.cn).}
\thanks{Masahiro Toyoura is with the School of Computer Science and Engineering, University of Yamanashi, Yamanashi 400-8510, Japan (e-mail: mtoyoura@yamanashi.ac.jp).}}

\maketitle

\begin{abstract}
A key challenge in the task of human pose and shape estimation is occlusion, including self-occlusions, object-human occlusions, and inter-person occlusions. The lack of diverse and accurate pose and shape training data becomes a major bottleneck, especially for scenes with occlusions in the wild. In this paper, we focus on the estimation of human pose and shape in the case of inter-person occlusions, while also handling object-human occlusions and self-occlusion. We propose a novel framework that synthesizes occlusion-aware silhouette and 2D keypoints data and directly regress to the SMPL pose and shape parameters. A neural 3D mesh renderer is exploited to enable silhouette supervision on the fly, which contributes to great improvements in shape estimation. In addition, keypoints-and-silhouette-driven training data in panoramic viewpoints are synthesized to compensate for the lack of viewpoint diversity in any existing dataset. Experimental results show that we are among the state-of-the-art on the 3DPW and 3DPW-Crowd datasets in terms of pose estimation accuracy. The proposed method evidently outperforms Mesh Transformer, 3DCrowdNet and ROMP in terms of shape estimation. Top performance is also achieved on SSP-3D in terms of shape prediction accuracy. Demo and code will be available at https://igame-lab.github.io/LASOR/.
\end{abstract}


\ifCLASSOPTIONpeerreview
\begin{center} \bfseries EDICS Category: 3-BBND \end{center}
\fi
%
\IEEEpeerreviewmaketitle

\section{Introduction}
%
%
%
%

 

\IEEEPARstart{H}{uman} pose and shape estimation from a single image is a challenging 3D vision problem. It may promote many promising applications, such as human-computer interaction, computer graphics, virtual fitting room, video understanding, etc. Recently   various deep-learning-based methods have been proposed to address this problem \cite{SMPL-X:2019,kolotouros2019spin,zanfir2018monocular,tan2017indirect,pavlakos2018learning}. Such methods provide impressive 3D human pose estimation results from a single RGB image. However, occlusion remains a major challenge for images in the wild, where self-occlusions, person-object occlusions and inter-person occlusions usually appear. It is difficult to achieve good performance when recovering the full 3D human body shape without explicitly taking occlusions into account. In this paper, we mainly focus on estimating the full 3D human pose and shape from a single RGB image under various occlusion cases.

\begin{figure}
    \centerline{\includegraphics[width=1.\linewidth]{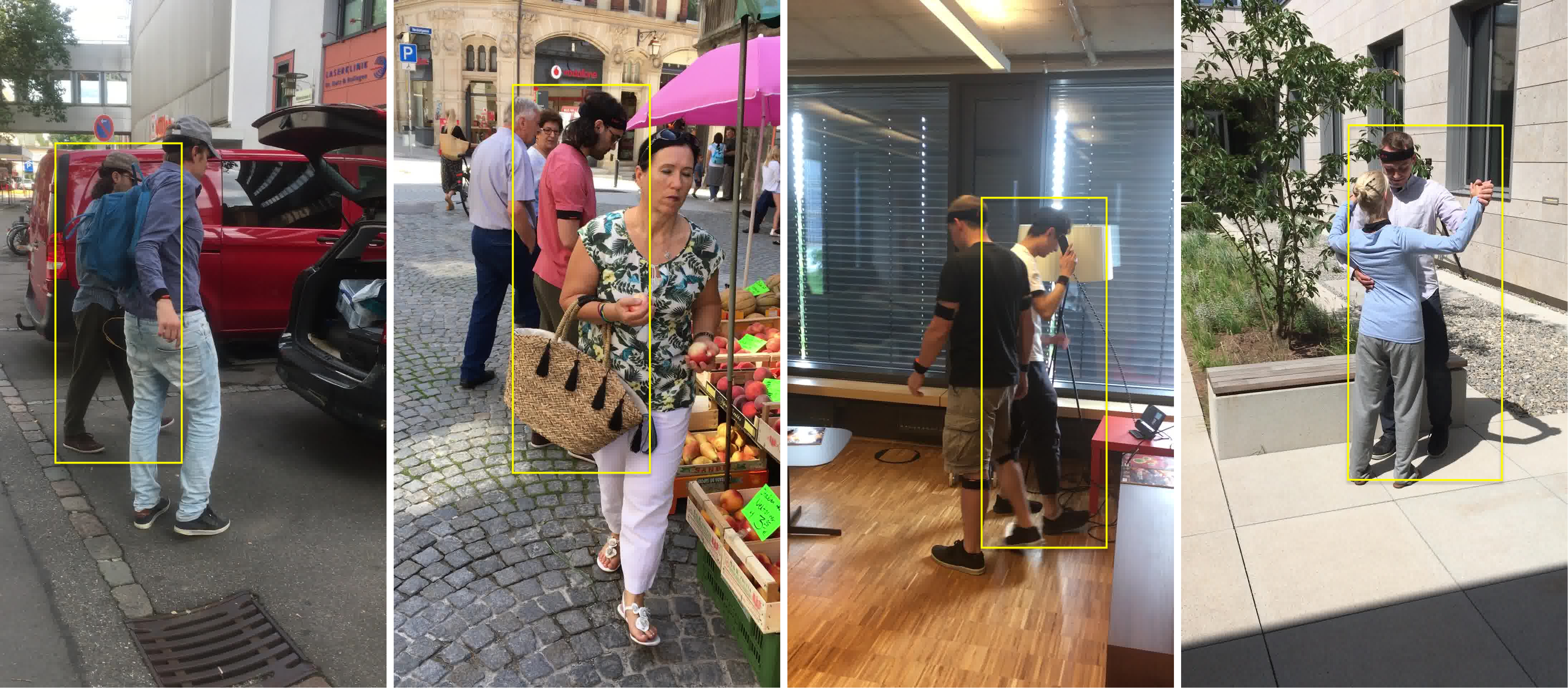}}
    \caption{Examples of inter-people occlusions. It is very challenging to estimate the pose and shape for the target persons enclosed by the yellow rectangles.}
    \label{fig:interpeople}
    \vspace{-0.2cm}
\end{figure}

By exploiting a parametric human body model SMPL \cite{SMPL:2015}, the task of human pose and shape estimation is usually converted into SMPL parameters estimation. SMPLify \cite{Bogo:ECCV:2016} fits the SMPL parameters to obtain a mesh consistent with image evidence, i.e., 2D keypoints. Omran et al. \cite{omran2018neural} regress the SMPL parameters with a Convolutional Neural Network. Kolotouros et al. \cite{kolotouros2019convolutional} directly regress the 3D locations of the mesh vertices using a Graph-CNN. With further development, some targeted questions have been raised. To estimate the shape more accurately, a method is proposed by synthesizing the input data with different shapes \cite{STRAPS2020BMVC}. For the object-occluded case, a method that utilizes a partial UV map to represent an object-occluded human body is proposed in \cite{zhang2020object} . Compared with the object-occluded cases, the person-occluded cases involve more complex scenes and yet are quite common in reality. Fig.\ref{fig:interpeople} shows some of such tricky cases. There are two main challenges. The first challenge is the lack of training data.  Existing datasets do not contain enough samples with occlusions. Moreover, since training data is traditionally collected by motion capture (MoCap) systems, and due to the limitations of existing MoCap systems, data collection of inter-person occlusions can hardly scale up. Another challenge is that severe inter-person occlusions would introduce challenging ambiguities into the network prediction, and thus confuse the full 3D human body shape estimation.

\begin{figure*}[htp]
    \centerline{\includegraphics[width=.9\linewidth]{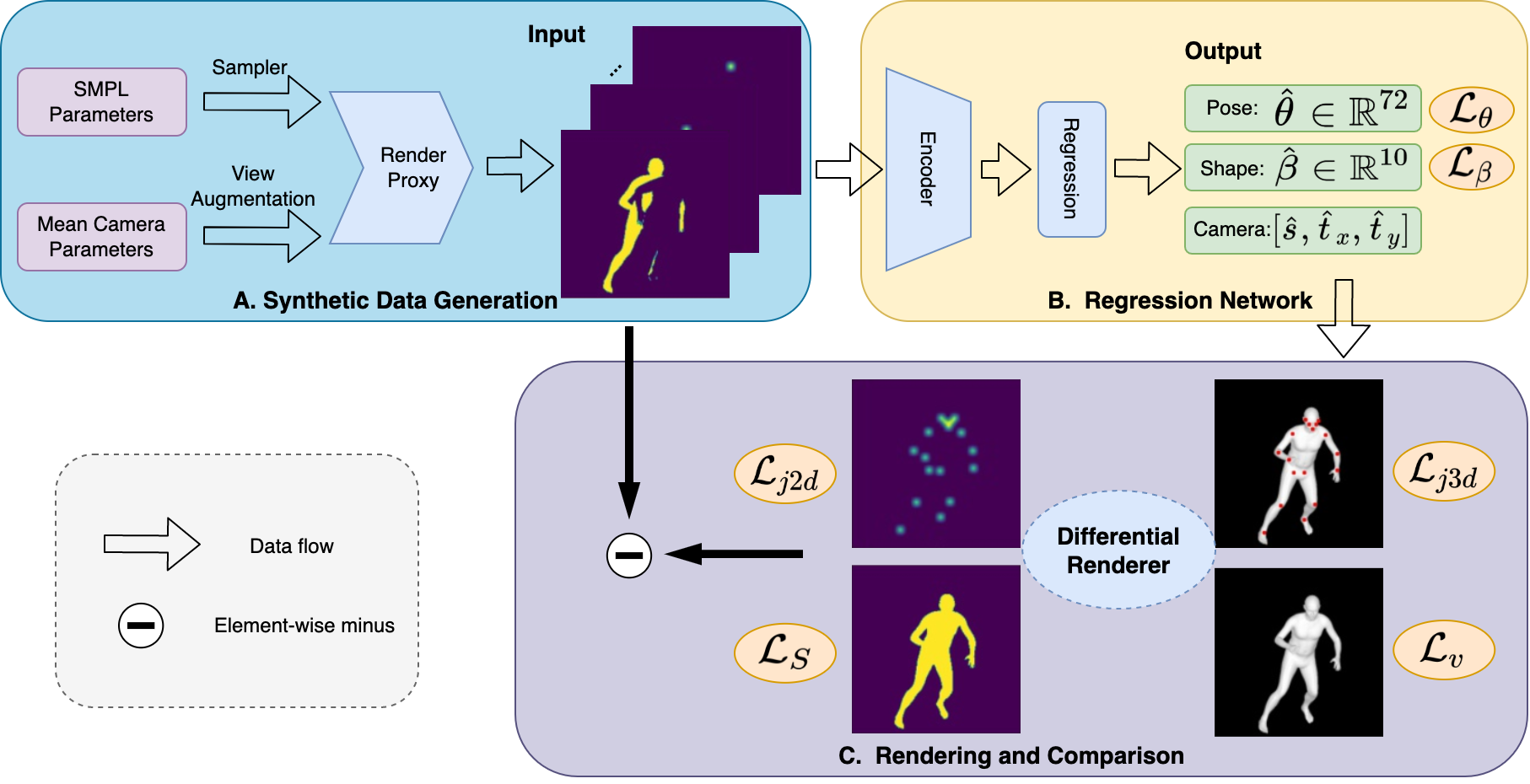}}
    \caption{\textbf{Overview of our method.} (A) is used to synthesize the training input. The input is then passed through the regressor network (B). We use rendering and compare the input with the rendered images in (C) to supervise the network.}
    \label{fig:overview}
\end{figure*}

To tackle the obstacles, inspired by \cite{STRAPS2020BMVC}, we propose to make use of synthetic keypoints-and-silhouette-driven training data and explicitly design an occlusion-aware framework during training. More specifically, silhouettes that reflect various inter-person occlusions are synthesized, and 2D keypoints are masked out if considered occluded. Using keypoints and silhouettes as the intermediate representation, large amount of inter-person occlusion samples can be generated with a low cost. Moreover, supervision of the silhouettes is enforced during training using a neural 3D mesh renderer integrated with the framework. The loss is backpropagated to the SMPL pose and shape parameters, which can enhance the human pose and shape estimation performance. Besides that, in many existing datasets, the cameras are installed in fixed positions. To further facilitate the network training and empower it with better generalization ability, we propose a viewpoint augmentation approach to create keypoints-and-silhouette-driven training data from panoramic views on top of any existing dataset. 

The overall flowchart of the proposed framework is shown in Fig.\ref{fig:overview}. Module A synthesizes the training data on-the-fly with the view augmentation and inter-person occlusions. The data then passes through the encoder and iterative regressor. The rendering and comparison structure is used to supervise the network. During inference, we just need the keypoint detector \cite{he2017mask} and DensePose \cite{guler2018densepose} to obtain the keypoints-and-silhouette data as input.

The main contributions of this work are summarized as follows:
\begin{itemize}
    \item We propose a robust occlusion-aware 3D pose and shape estimation framework that exploits synthetic keypoints-and-silhouette-driven training data, which overcomes the data scarcity especially for person-occluded scenes in the wild.

    \item Neural 3D mesh renderer is integrated with the framework to provide silhouette supervision during training, which contributes to much more accurate shape estimation.

    \item A viewpoint augmentation strategy is introduced to synthesize infinite viewpoint variation, creating keypoints-and-silhouette-driven training data from panoramic views on top of any existing dataset.

\end{itemize}

\section{Related Work}
\IEEEPARstart{I}{n} this section, we discuss recent 3D human pose and shape estimation approaches, as well as the related methods for the occlusion problem.

\noindent\textbf{Optimization-based methods.} Optimization-based approa-ches attempt to fit a parametric body model, like SMPL \cite{SMPL:2015} or SCAPE \cite{anguelov2005scape}, to 2D observations. SMPLify \cite{Bogo:ECCV:2016} is the first method to automatically estimate 3D pose and shape from a single unconstrained image by fitting the SMPL to 2D keypoint detections. Other than 2D keypoint detections, different cues like body surface landmarks \cite{lassner2017unite}, silhouettes \cite{lassner2017unite} or body part segmentation \cite{zanfir2018monocular} are also used in the fitting procedure. Recent works try to fit a more expressive model \cite{joo2018total,xiang2019monocular} that includes face, body, and hand details. Gu et al. \cite{gu2019multi} adopt a hierarchical 3D human model that imposes angle and bone length constraints for multi-person pose estimation. These approaches produce reliable results with good 2D observations. However, they are sensitive to initialisation and rely heavily on the quality of 2D information.

\noindent\textbf{Regression-based methods.} To this end, recent works rely almost exclusively on deep neural networks to regress the 3D human pose and shape parameters. Owing to the lack of training data with full 3D shape ground truth, 2D annotations including 2D keypoints, silhouettes, or parts segmentation are usually essential. This information can be used as intermediate representation \cite{kato2018neural,xu2019denserac}, or as supervision \cite{pavlakos2018learning,kato2018neural}. Kolotouros et al. \cite{kolotouros2019convolutional} directly regress the 3D location of the mesh vertices from a single image instead of predicting model parameters using a Graph-CNN. Pose2mesh \cite{Choi_2020_ECCV_Pose2Mesh} takes a 2D pose as input and avoids the domain gap between training and testing. I2L-MeshNet \cite{Moon_2020_ECCV_I2L-MeshNet} predicts the
per-lixel (line+pixel) likelihood on 1D heatmaps for each mesh vertex coordinate instead of directly regressing the parameters. Fundamentally, regression-based methods are dependent on the sample diversity of the training data. However, since obtaining 3D training data is awfully expensive, most datasets only contain  a limited number of human bodies. This is also the reason why the existing datasets are insufficient to support robust and accurate human body shape estimation in the wild.

\noindent\textbf{Methods for occlusion problems.} Huang and Yang \cite{huang2009estimating} propose a method to recover 3D pose when a person is partially or severely occluded in the scene from monocular images. However, the occlusions are limited to simple rectangles. A grammar-based model is introduced by \cite{girshick2011object} with explicit occlusion part templates. To avoid specific design for occlusion patterns. Ghiasi et al. \cite{ghiasi2014parsing} present a method to model occlusion by learning deformable models with many local part mixture templates using large quantities of synthetically generated training data, which is aimed at explicitly learning the appearance and statistics of occlusion patterns. Zhang et al. \cite{zhang2020object} present a method that utilizes a partial UV map to represent an object-occluded human body, and the full 3D human shape estimation is ultimately converted as an image inpainting problem. The above methods mainly deal with the object-human occlusion problems. Bin et al. \cite{bin2020adversarial} present a method that augments images by pasting segmented body parts with various semantic granularity, which has certain generalization ability to shield the human body from each other. The occlusion problem is explored in \cite{gu2021exploring} by using a temporal gated convolution network. Choi et al. \cite{lin2021end-to-end} present 3DCrowdNet, a 2D human pose-guided 3D crowd pose and shape estimation system for in-the-wild scenes. Sun et al. \cite{ROMP} develop a novel Collision-Aware Representation (CAR) method named ROMP to improve the performance under person-person occlusion. Jiang et al. \cite{jiang2020coherent} propose a depth ordering-aware loss to generate a rendering that is consistent with the annotated instance segmentation, but they make use of very few datasets with 3D ground truth. Currently, occlusion is still a major challenge for 3D pose estimation in the wild.  

\noindent\textbf{Synthetic training data.} When the data is too scarce to solve the problem, synthetic data becomes an option. Several papers explore the use of synthetic data in training for human pose and shape estimation. To make up for the scarcity of in-the-wild training data with diverse and accurate body shape labels, Sengupta et al. \cite{STRAPS2020BMVC} propose STRAPS, a system that utilizes proxy representations, such as silhouettes and 2D joints, as inputs to a shape and pose regression neural network, which is trained with synthetic training data. Some methods \cite{bin2020adversarial,zhang2020object} try to obtain novel samples either by adding virtual objects to existing datasets or by pasting segmented body parts with various semantic granularity. However, while inter-person occlusion is common in 3D human pose estimation scenarios, the inter-person occlusion is rarely synthesized explicitly. 

\begin{figure*}[!htp]
    \centerline{\includegraphics[width=0.9\linewidth]{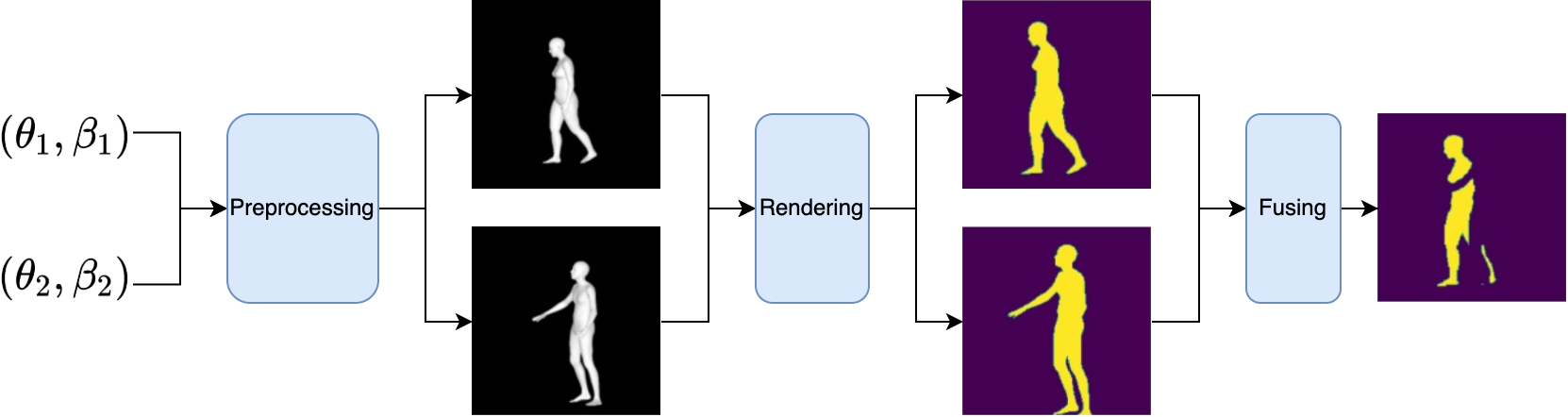}}
    \caption{\textbf{Synthesizing the silhouette.} First, two pairs of SMPL parameters $(\bm{\theta}_1, \bm{\beta}_1)$,$(\bm{\theta}_2, \bm{\beta}_2)$ are sampled. After viewpoint and shape augmentation is performed, a neural mesh renderer generates the 2D silhouettes. Finally, the pair of silhouettes are used to synthesize the overlap of human body.}
    \label{fig:syn_input}
\end{figure*}

\quad 

\section{Technical approach}
\IEEEPARstart{I}{n} the following section, we first describe the Skinned Multi-Person Linear model (SMPL) briefly and the basic notations defined in this paper. More details about the data generation are then provided. After that, the regression network and the loss function are presented in detail.

\subsection{SMPL Model}
 SMPL is a skinned vertex-based model that accurately represents a wide variety of body shapes in natural human poses. SMPL model provides the function $\mathcal{M}(\bm{\theta}, \bm{\beta})$ that takes as input the pose parameters $\bm{\theta}$ and the shape parameters $\bm{\beta}$ and returns the body mesh $\bm{v} \in \mathbb{R}^{N \times 3}$ with $N=6890$ vertices. The 3D joints are calculated by $\bm{j}^{3d}=\bm{J}\bm{v}$, where $\bm{J}\in \mathbb{R}^{K \times N}$ is a pre-trained linear regressor matrix, and $K$ is the number of joints.  
\subsection{View Augmentation}

For the datasets collected in the laboratory, the camera is usually placed in a fixed position. Meanwhile, datasets collected outdoors with 3D ground truth are scarce and of insufficient variations as well. Models trained using these datasets may depend on the camera's perspective and have poor generalization ability. To this end, this paper adopts a viewpoint augmentation strategy to generate intermediate representations, i.e., keypoints and silhouettes, under different camera parameters. Using these intermediate representations can make full use of this feature and reduce the model's dependence on a specific camera perspective. In the implementation, since the first three values of the SMPL model pose parameters control the global orientation of the human body model, we alter these parameters to modify the current human body orientation relative to the camera. Modifying the global orientation of the human body model does not modify the 3D posture of the human body, but the projection on the 2D image is quite different. The keypoints and contour maps of the input model in this paper are all two-dimensional data. This greatly enriches the training data when the camera viewpoint in the training data is limited. The specific formula is given as follows:
\begin{equation}
    \hat{\bm{\theta}}_{g} = \bm{\theta}_{g} + \pi*\mathcal{N}(\mu, \sigma^{2}) 
\end{equation}
where $\bm{\theta}_{g}$ is the global orientation in the $\bm{\theta}$ parameter, and $\mathcal{N}$ is  a normal distribution function with mean $\mu$ and standard derivation $\sigma$.

\subsection{Synthetic Data Generation}
\begin{figure}[htbp]
    \centerline{\includegraphics[width=.9\linewidth]{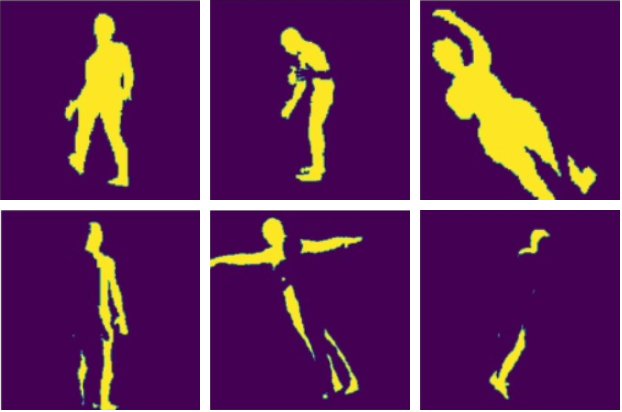}}
    \caption{Examples of synthetic data generation with different levels of inter-person occlusions. }
    \label{fig:samples}
\end{figure}

As shown in Fig. \ref{fig:syn_input}, two pairs of SMPL parameters $(\bm{\theta}_1, \bm{\beta}_1)$,$(\bm{\theta}_2, \bm{\beta}_2)$ are sampled from any training dataset. The camera translation vector $\bm{t}_1$ is generate randomly, while the camera intrinsic matrix $K$ and rotation matrix $R$ are fixed. Then, we obtain $\bm{t}_2$ based on $\bm{t}_1$ by adding a shift. In order to get various shape data, we replace $\bm{\beta}_{i}$ with a new random vector $\bm{\beta}^{'}$, generated by sampling $n$ shape parameters $\bm{\beta}_{n}^{'} \sim \mathcal{N}(\mu,\sigma_{n}^{2})$ \cite{STRAPS2020BMVC}. We use $\bm{\theta}_{1}^{'}$ and $\bm{\theta}_{2}^{'}$ by replacing the global orientation with the orientation $\hat{\bm{\theta}}_{g}$ obtained by viewpoint augmentation method instead of the original $\bm{\theta}_{1}$ and $\bm{\theta}_{2}$. Then, the 3D vertices $\bm{v}_1$ and $\bm{v}_2$ corresponding to $(\bm{\theta}_{1}^{'}, \bm{\beta}_1^{'})$ and $(\bm{\theta}_{2}^{'}, \bm{\beta}_2^{'})$ are rendered \cite{kato2018neural} into silhouette $S_{1},S_{2} \in [0, 1]^{H \times W}$. The final silhouette is calculated by {\eqref{equ:sil}} as follows,
\begin{equation}
    S = S_{1}-S_{1}*S_{2}
    \label{equ:sil}
\end{equation}
As shown in Fig. \ref{fig:samples}, inter-person occlusion samples are generated with different levels of human body overlap.

For 2D joints, the 3D joints $\bm{j}^{3d}$ corresponding to $(\bm{\theta}_1^{'}, \bm{\beta}_1^{'})$ are projected into 2D joints $\bm{j}^{2d} \in \mathbb{R}^{K \times 2}$ using prospective projection. $\bm{j}^{2d}$ is encoded into 2D Gaussian joint heatmaps, $G \in \mathbb{R}^{K\times H\times W}$. We obtain the final input $X \in \mathbb{R}^{(K+1)\times H \times W}$ by concatenating S and G. Note that the noises in keypoints detection will introduce noises into the 2D joint heatmap encoding. To handle occluded joints explicitly, we randomly remove some body parts from the human body and add occluding boxes to the silhouette in $S$ during training.

\subsection{Regression Network}
\label{sec:regression}
Our method is architecture-agnostic. We use the baseline network architecture as in \cite{ kanazawa2018end, kolotouros2019spin,STRAPS2020BMVC}, which consists of a convolutional encoder for feature extraction and an iterative regressor that outputs predicted SMPL pose, shape and camera parameters ($\hat{\bm{\theta}}$, $\hat{\bm{\beta}}$ and $\hat{\bm{\Pi}}$) from the extracted feature. Similarly as in STRAPS \cite{STRAPS2020BMVC}, the SMPL default discontinuous Euler rotation vectors are replaced by a continuous 6-dimensional rotation representation of $\hat{\bm{\theta}}$ proposed by \cite{zhou2019continuity}.

The weak-perspective camera model is adopted in this paper, represented by $\hat{\bm{\Pi}} = [\hat{s},\hat{\bm{t}}]$ where $\hat{s} \in \mathcal{R}$ means scale and $\hat{\bm{t}} \in \mathcal{R}^2$ represents ${x-y}$ camera translation. These parameters allow us to generate the mesh corresponding to the regressed parameters, $\hat{\bm{v}}=\mathcal{M}(\hat{\bm{\theta}}, \hat{\bm{\beta}})$, as well as the 3D joints $\hat{\bm{j}}^{3d} = \bm{J}\hat{\bm{v}}$ and their 2D reprojection $\hat{\bm{j}}^{2d}=\hat{\bm{\Pi}}(\hat{\bm{j}}^{3d})$.

\noindent\textbf{Neural 3D Mesh Renderer.} Differentiable rendering is a novel field which allows the gradients of 3D objects to be calculated and propagated through images. We follow the common self-supervision pipeline with differentiable rendering and don't need correspondence to images. A neural 3D mesh renderer  \cite{kato2018neural} is integrated into our framework to provide silhouette supervision during training. The loss function is defined as 
\begin{equation}\label{eq_L_sil}
    \mathcal{L} _{S} = \frac{1}{H\times {W}}\times \left \| \hat{S} - S  \right \|  _{2}^{2}
\end{equation}
where $S$ is the target silhouette image, $\hat{S}$ is silhouette rendered from $\hat{\bm{v}}$ and H, W are the height and width of the target silhouette $S$. The reconstruction of 3D shape is an ill-posed problem when a single view is used during training. Additional loss functions are added to provide strong supervision.

The first loss term $\mathcal{L}_v$ performs the supervision between predicted 3D vertices $\hat{\bm{v}}$ and ground-truth 3D vertices $\bm{v}$, which is,
\begin{equation}
    \mathcal{L}_v = \left \| \hat{\bm{v}} -\bm{v}   \right \| _{2}^{2}
\end{equation}

The second loss term $\mathcal{L}_{j3d}$ performs the supervision between predicted 3D joints and ground-truth 3D joints, which is,
\begin{equation}
    \mathcal{L}_{j3d} = \left \| \hat{\bm{j}}^{3d} - \bm{j}^{3d}  \right \|  _{2}^{2}
\end{equation}
The $\mathcal{L}_{\theta}$ and $\mathcal{L}_{\beta}$ losses can provide strong 3d supervision.
\begin{gather}
    \mathcal{L}_{\theta} = \left \| \hat{\theta} - \theta  \right \|  _{2}^{2} \\
    \mathcal{L}_{\beta} = \left \| \hat{\beta} - \beta  \right \|  _{2}^{2}
\end{gather}

In order to enforce image-model alignment, we add the loss for the projected 2D joints
\begin{equation}
    \mathcal{L}_{j_{2d}}=\frac{1}{K} \times {\textstyle \sum_{i=1}^{K}} \omega _{i}\left \| \hat{\bm{j}}^{2d}_{i}-\bm{j}^{2d}_{i}  \right \| _{2}^{2}
\end{equation}
where the value of $\omega_{i}$ indicates the visibility (1 if visible, 0 otherwise) for i-th 2D keypoints (K in total).
The final loss function can be described as follows:
\begin{equation}
\begin{split}
    \mathcal{L}=\frac{1}{\sigma_{v}^{2}} \mathcal{L}_{v}+\frac{1}{\sigma_{j_{3d}}^{2}} \mathcal{L}_{j_{3d}}+\frac{1}{\sigma_{\theta}^{2}} \mathcal{L}_{\theta}+\frac{1}{\sigma_{\beta}^{2}} \mathcal{L}_{\beta}+\frac{1}{\sigma_{j_{2d}}^{2}} \mathcal{L}_{j_{2d}} \\
    +\frac{1}{\sigma_{S}^{2}} \mathcal{L}_{S}
    +\log \left(\sigma_{v} \sigma_{j_{3d}} \sigma_{\theta} \sigma_{\beta} \sigma_{j_{2d}} \sigma_{S}\right)
\end{split}
\end{equation}
where losses are adaptively combined using homoscedastic uncertainty \cite{kendall2018multi}.
In the next section, we will analyze the contribution of the silhouette loss $\mathcal{L}_{S}$ in 3D shape estimation accuracy.

\section{Experiments}
\begin{figure*}[htbp]
    
    \centerline{\includegraphics[width=0.75\linewidth]{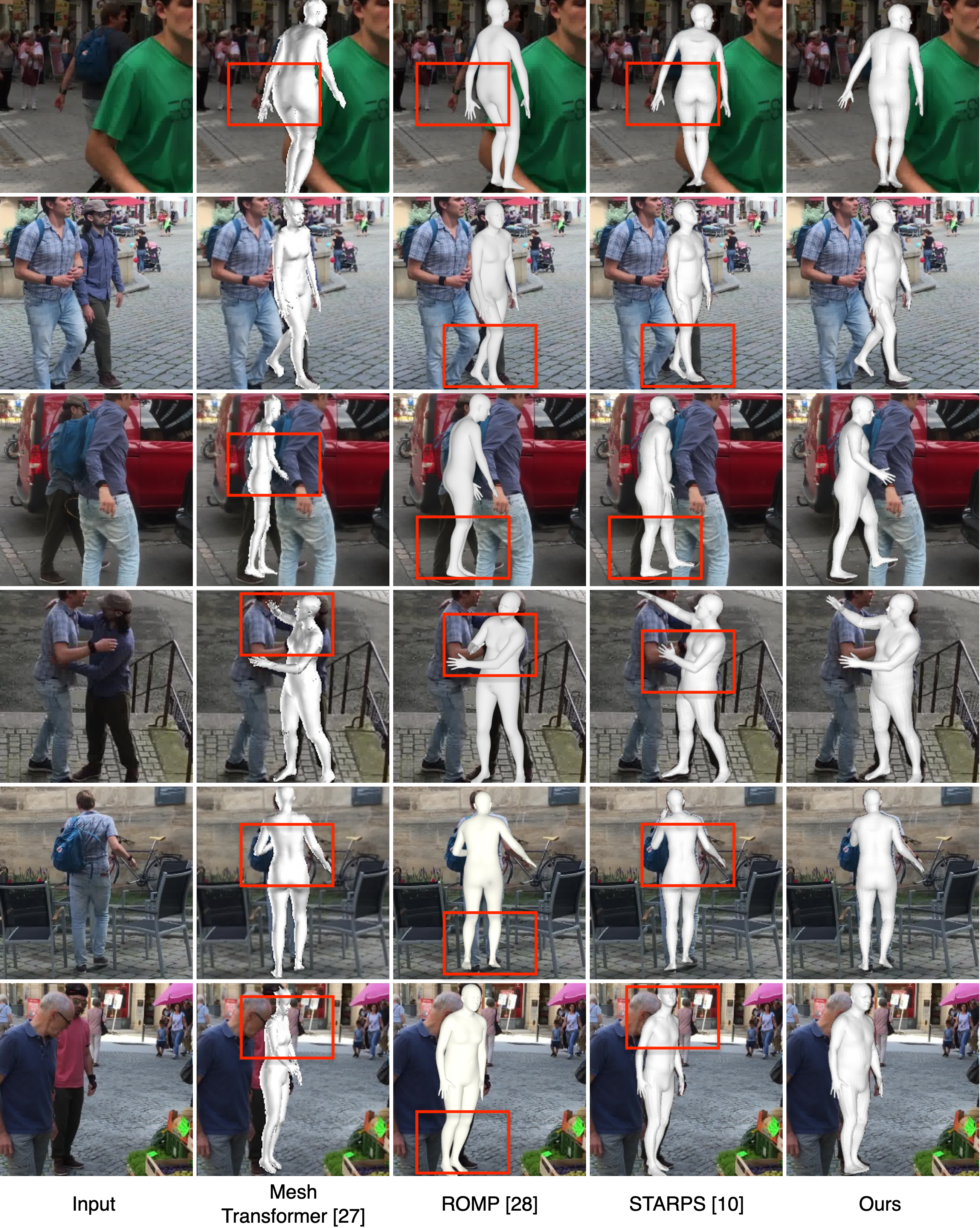}}
    \caption{\textbf{Qualitative comparison with the state-of-the-art methods.} Some samples with different levels of inter-person occlusions from 3DPW dataset are shown. Compared with Mesh Transformer \cite{lin2021end-to-end}, ROMP \cite{ROMP} and STRAPS \cite{STRAPS2020BMVC}, our method gives more accurate pose and shape predictions for the person-occluded cases.}
    \label{fig:people-occluded}
\end{figure*}
\subsection{Datasets}
This section gives a description of the datasets used for training and evaluation.
Following STRAPS\cite{STRAPS2020BMVC}, we train the network using synthetic data generated from 3DPW \cite{vonMarcard2018}, AMASS \cite{AMASS:ICCV:2019}, and UP-3D \cite{lassner2017unite}. We report evaluation results on 3DPW, 3DOH50K \cite{zhang2020object} and SSP-3D \cite{STRAPS2020BMVC}.

\textbf{AMASS} is a large and varied database of human motion that unifies 15 different optical marker-based mocap datasets by representing them within a common framework and parameterizaition. Some of the SMPL parameters are selected to train our network.

\textbf{UP-3D} combines two LSP datasets and the single-person part of MPII-HumanPose dataset, which is used to train our network.

\textbf{3DPW} is captured via IMUs and contains both indoor and outdoor scenes. It provides abundant 2D/3D annotations, such as 2D pose, 3D pose, SMPL parameters, human 3D mesh, etc. It includes inter-person occlusion, object-human occlusion, and non-occluded/truncated cases. Being a rare in-the-wild 3D human pose and shape dataset, 3DPW is the most suitable for testing our work. 3DPW is used for training and evaluation with the same split as in \cite{STRAPS2020BMVC}.

\textbf{3DPW-Crowd} is a subset of the 3DPW validation set and the selected sequences are courtyard\_hug\_00 and courtyard\_dancing\_00. 3DPW-Crowd is deemed a suitable benchmark for evaluating performance for inter-person occlusion scenes in \cite{choi20213dcrowdnet}, as it has much higher bounding box IoU and CrowdIndex\cite{li2019crowdpose}, which measures crowding level.

\textbf{SSP-3D} contains 311 in-the-wild images of 62 tightly clothed sports persons (selected from the Sport-1M video dataset) with a diverse range of body shapes, along with the corresponding pseudo-ground-truth SMPL shape and pose labels. 

\textbf{3DOH50K} is the first 3D human pose dataset that introduces object-human occlusions on purpose when recording. It contains 1290 test images. This dataset is used for evaluation.

\begin{table}[htbp]
\caption{Comparison of different evaluation datasets related to 3D pose estimation. The Object-Occluded information is provided by OOH \cite{zhang2020object}, where the + mark(s) indicates the level of occlusions.}
\begin{center}
\begin{tabular}{@{}lccc@{}}
\toprule
Dataset   & Object-Occluded & Inter-person Occlusion & Num of People \\ \midrule
3DPW      & ++              & \CheckmarkBold           & 18            \\
3DPW-Crowd    & -               & \CheckmarkBold             & 2         \\
SSP-3D    & -               & \XSolidBrush             & 62         \\
3DOH50K   & ++++            & \XSolidBrush              & -      \\ \bottomrule
\end{tabular}
\label{datasets analysis}
\end{center}
\end{table}

\begin{table*}[htbp]
\caption{\textbf{Comparisons with the state-of-the-art methods on 3DPW, 3DPW-Crowd, 3DOH50K and SSP-3D.} Number marked with * were evaluated using the open-sourced official code. Number marked with - were provided by \cite{STRAPS2020BMVC}, all other numbers are reported by the respective papers. In order to be more concise, the best performance given by methods without using the paired ground-truth 3D labels and images during training are marked in \textbf{bold black}, and other methods are marked in {\color[HTML]{32CB16}light green}.}
\begin{center}
\begin{tabular}{lccccc} 
\toprule
\multirow{2}{*}{Method}      & 3DPW          & 3DPW-Crowd & 3DOH50K   & \multicolumn{2}{c}{SSP-3D}  \\
                             & MPJPE-PA              & MPJPE-PA           & MPJPE-PA           & PVE-T-SC                          &mIOU$\uparrow$                               \\ 
\hline
STRAPS \cite{STRAPS2020BMVC}                     & 66.8                  & $70.5^{*}$            & $73.5^{*}$              & 15.9                              & \textbf{0.8}                       \\
HMR(unpaired) \cite{kanazawa2018end}            & -                     & -                  & -                  & $20.8^{-}$                        & $0.61^{-}$                         \\
Pose2Mesh \cite{Choi_2020_ECCV_Pose2Mesh}   & 58.3 &79.8               & - & -                & -                 \\
Ours                         & \textbf{57.9}         & \textbf{67.6}     & \textbf{72.5}      & \textbf{14.5}                     & 0.67                               \\ 
\hline
HMR   \cite{kanazawa2018end}                       & 76.7                  & -                  & 92.9               & $22.9^{-}$                        & $0.69^{-}$                         \\
SPIN \cite{kolotouros2019spin}                        & 59.2                  & 69.9               & 73.6               & ${\color[HTML]{32CB16} 22.2}^{-}$ & ${\color[HTML]{32CB16} 0.70}^{-}$  \\
ROMP   \cite{ROMP}                      & 54.9                  & -                  & ${\color[HTML]{32CB16} 43.9}$               & -                                 & -                                  \\

I2L-MeshNet \cite{Moon_2020_ECCV_I2L-MeshNet} & 57.7 & 73.5               & - & -               & -                 \\
3DCrowdNet \cite{choi20213dcrowdnet}  & 52.2 &  ${\color[HTML]{32CB16} 56.8}$             &- & -                & -               \\
Mesh Transformer \cite{lin2021end-to-end}  &${\color[HTML]{32CB16} 47.9}$ & -               & - & -                & -              \\
OOH\cite{zhang2020object}                          & 72.2                  & -                                   & 58.5               & -                                 & -                                  \\
\bottomrule
\end{tabular}

\label{table:total}
\end{center}
\end{table*}

\noindent\textbf{Dataset Statistics.} Our work mainly focuses on human pose and shape estimation with inter-person occlusions. We compare different datasets to give insights about the types of occlusion and the number of people in Table \ref{datasets analysis}.

\subsection{Implementation Details}
\noindent\textbf{Network Architecture.} For a fair comparison with previous methods, we use fine-tuned  ResNet-18 \cite{he2016deep} as the default encoder and ResNet-50 as the additional encoder. The encoder produces a feature vector $\phi \in \mathbb{R}^{512}$. The iterative regression network consists of two fully connected layers. During inference, we use darkpose \cite{Zhang_2020_CVPR} to get 2D keypoints of 3DPW and use Mask-RCNN \cite{he2017mask} to get 2D keypoint predictions of other datasets. We use DensePose \cite{guler2018densepose} to get silhouette predictions. All implementations are done in PyTorch \cite{paszke2019pytorch}. 

\noindent\textbf{Setting Details.} The input silhouettes are cropped and resized to $256\times256$, while keeping the aspect ratio of 1.2. COCO keypoint order is used and the number of keypoint is 17. The input tensor is $18\times256\times256$. The loss weights are set to be  $\sigma_{v}=1.0, \sigma_{j_{3d}}=1.0, \sigma_{\theta}=0.1, \sigma_{\beta}=0.1, \sigma_{j_{2d}}=0.1, \sigma_{S}=0.1$. We use the Adam optimizer \cite{kingma2014adam} with a batch size of 140 and an initial learning rate of 0.0001 to train our encoder and regressor. The mean values in iterative regression are the same as SPIN \cite{kolotouros2019spin}. The training step takes 10 days with 100 epochs on a single 3090 GPU.

\noindent\textbf{Training Datasets.} For a fair comparison with the previous method STRAPS \cite{STRAPS2020BMVC}, we use the training data provided by the STRAPS \cite{STRAPS2020BMVC}. The training data are collected from AMASS, 3DPW and UP-3D. In the training step, we use the ground-truth SMPL parameters and do not use any pair of image and 3D label.

\noindent\textbf{Evaluation Metrics.} We adopt Procrustes-aligned \cite{mehta2017monocular} mean per joint position error (MPJPE\_PA) for evaluating the 3D pose accuracy. To evaluate the 3D shape error, we employ Procrustes-aligned per-vertex error (PVE\_PA). Besides, to evaluate the shape estimation, we calculate the scale-corrected per-vertex euclidean error in a neutral pose (PVE-T-SC) and mean intersection-over-union (mIOU) on SSP-3D. In all the tables, $\uparrow$ means the larger the better, otherwise it is the lower the better.

\subsection{Comparisons to the State-of-the-Art:}

To demonstrate the effectiveness of our method, we perform quantitative evaluations on 3DPW, 3DOH50K and SSP-3D. In Table \ref{table:total}, the methods in the upper rows do not use any paired images and 3D labels. Separated by the horizontal line, the methods in the bottom rows use the paired ground-truth 3D labels and images during training. Our method does not directly use any real images in the training step.

\begin{figure*}[htbp]
    
    \centerline{\includegraphics[width=.9\linewidth]{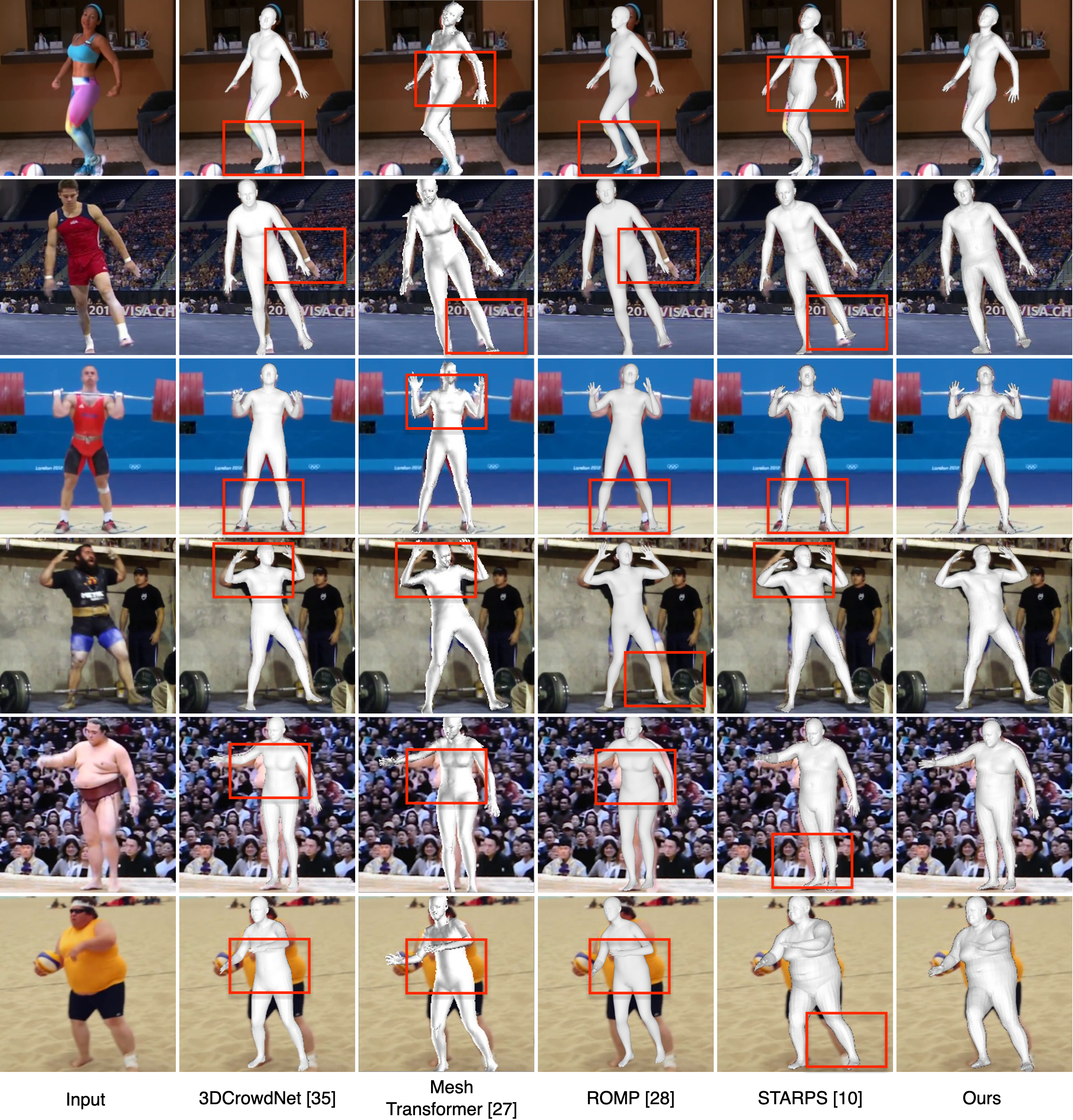}}
    \caption{\textbf{Qualitative comparison for different body shapes with the state-of-the-art methods 3DCrowdNet \cite{choi20213dcrowdnet}, Mesh Transformer \cite{lin2021end-to-end}, ROMP \cite{ROMP} and STRAPS \cite{STRAPS2020BMVC}.} The samples are from the SSP-3D dataset. The results of ROMP \cite{ROMP}, STRAPS \cite{STRAPS2020BMVC}, and Mesh Transformer \cite{lin2021end-to-end} are generated from their corresponding official codes, and the results of 3DCrowdNet \cite{choi20213dcrowdnet} are provided by  the author. Our method is able to accurately predict a diverse range of body shapes.}
    \label{fig:ssp}
\end{figure*}

\noindent\textbf{3DPW.} Firstly, we evaluate the test dataset in 3DPW. As shown in Table \ref{table:total}, our method achieves a 12 percent improvement compared with \cite{STRAPS2020BMVC}. Although our method does not use paired images and 3D labels, it shows competitive performance with the state-of-the-art that requires training data comprised of paired images and 3D labels. Imposing less restrictions on the training data means our method can scale up easier without relying on paired images and labels. Our method also shows excellent shape estimation performance.  Note that while Mesh Transformer \cite{lin2021end-to-end}, 3DCrowdNet \cite{choi20213dcrowdnet} and ROMP \cite{ROMP} report higher MPJPE-PA accuracy on 3DPW, the proposed method clearly shows better performance on the silhouette for human with various shapes, as shown in Fig. \ref{fig:ssp}.

\noindent\textbf{3DPW-Crowd.} We also evaluate the proposed method for 3DPW-Crowd. As shown in Table \ref{table:total}, our method achieves the best performance among the methods that do not need paired training data (top rows), including STRAPS \cite{STRAPS2020BMVC} and Pose2Mesh \cite{Choi_2020_ECCV_Pose2Mesh}. 

\noindent\textbf{SSP-3D.} The SSP-3D dataset contains 62 subjects. The proposed method shows the top performance on shape estimation in terms of PVE-T-SC.

\noindent\textbf{3DOH50K.} In 3DOH50K dataset, the samples have severe occlusions. As shown in Table \ref{table:total}, better performance is achieved compared with \cite{STRAPS2020BMVC} on handling object-human occlusions. The other methods also use the Human3.6M dataset to train the model while ours does not. Therefore, it would not be fair to directly compare our results to theirs.

Finally, to show the effectiveness of our method on handling inter-person occlusions, some examples are also shown in Fig. \ref{fig:people-occluded}. To compare the shape estimation of the human body more intuitively, some examples with diverse body shapes are demonstrated in Fig. \ref{fig:ssp}.

\subsection{Ablation Study}
\begin{table}[htbp]
\caption{ Evaluation on the SSP-3D dataset.}
\begin{center}
\begin{tabular}{@{}lccc@{}}
\toprule
Method          & PVE-T-SC & mIOU$\uparrow$ & convergence time (epochs) \\ \midrule
W/O silhouette loss   & 17.2  & \textbf{0.67}  & 250 \\
W/ silhouette loss & \textbf{14.5}& \textbf{0.67}  & \textbf{100}\\ \bottomrule
\end{tabular}
\label{silh}
\end{center}
\end{table}

\begin{figure}[htbp]
    \centerline{\includegraphics[width=0.9\linewidth]{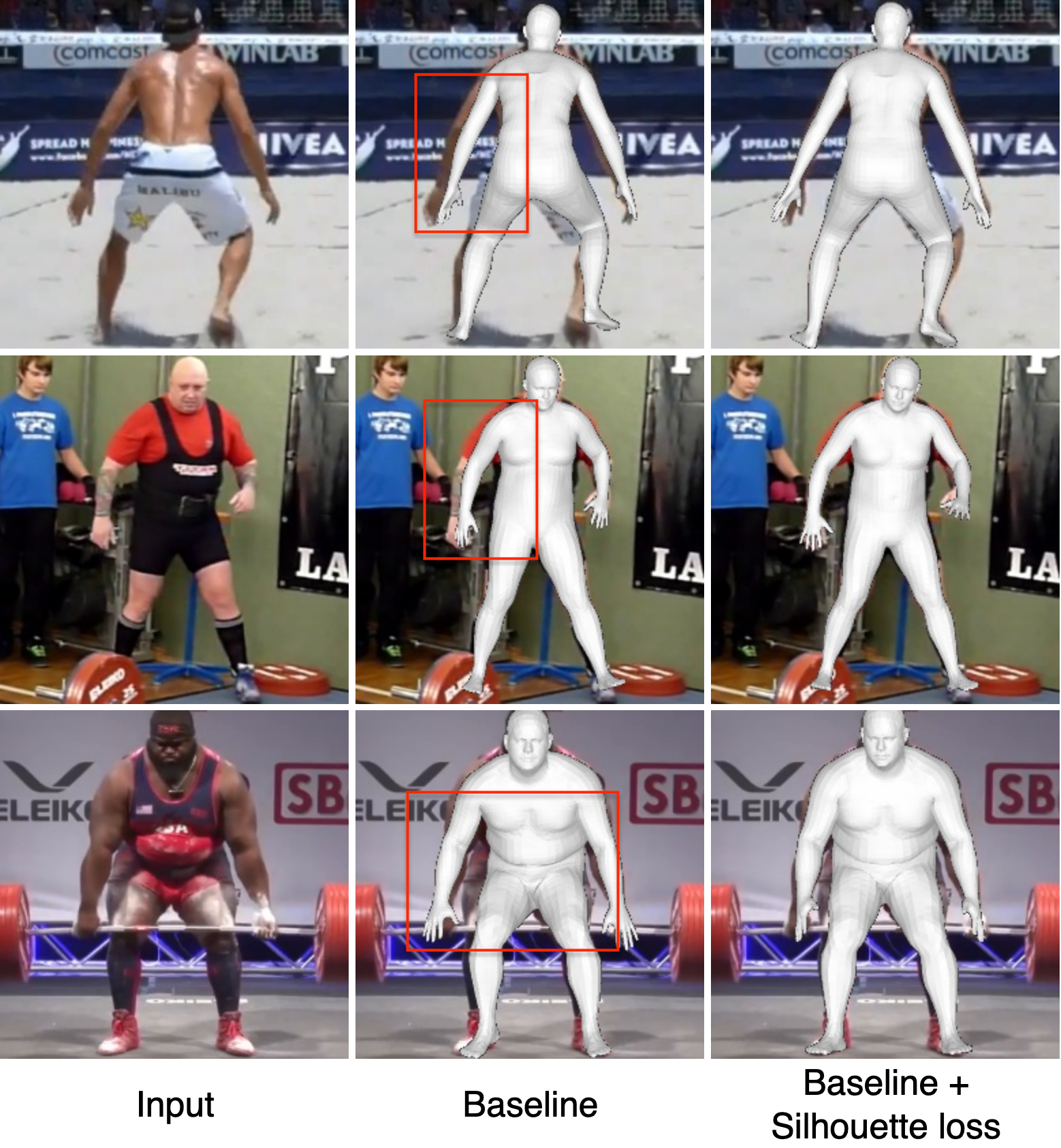}}
    \caption{Ablation study of the silhouette loss enabled by neural 3D mesh rendering.}
    \label{fig:sil}
\end{figure}

\noindent\textbf{Silhouette Loss.} To study the effect of the neural-mesh-rendering-enabled silhouette loss $\mathcal{L}_{S}$ in {\eqref{eq_L_sil}}, we compare two models trained (i) without the silhouette loss, and (ii) with the silhouette loss. The results are shown in Table \ref{silh}. The qualitative comparison with the baseline and silhouette loss are shown in Fig. \ref{fig:sil}. The silhouette loss improves both the network convergence and the final performance. For the model with silhouette supervision, training converges within 100 epochs, whereas the baseline model takes 250 epochs. Note that the baseline uses the Human3.6m dataset while our method does not use it. Our method still gives better shape estimation in terms of PVE-T-SC.

\noindent\textbf{Silhouette As Intermediate Representation.} To study the impact of having silhouette as intermediate representation, we compare two models trained with (i) keypoint only, and (ii) keypoint and silhouette, both with the process of synthetic data generation. The results are shown in Table \ref{tab:ks}. The method with silhouette as intermediate representation consistently shows a higher performance. This proves that the combination of keypoints and silhouette representations is necessary.

\begin{figure}
    
    \centerline{\includegraphics[width=1.0\linewidth]{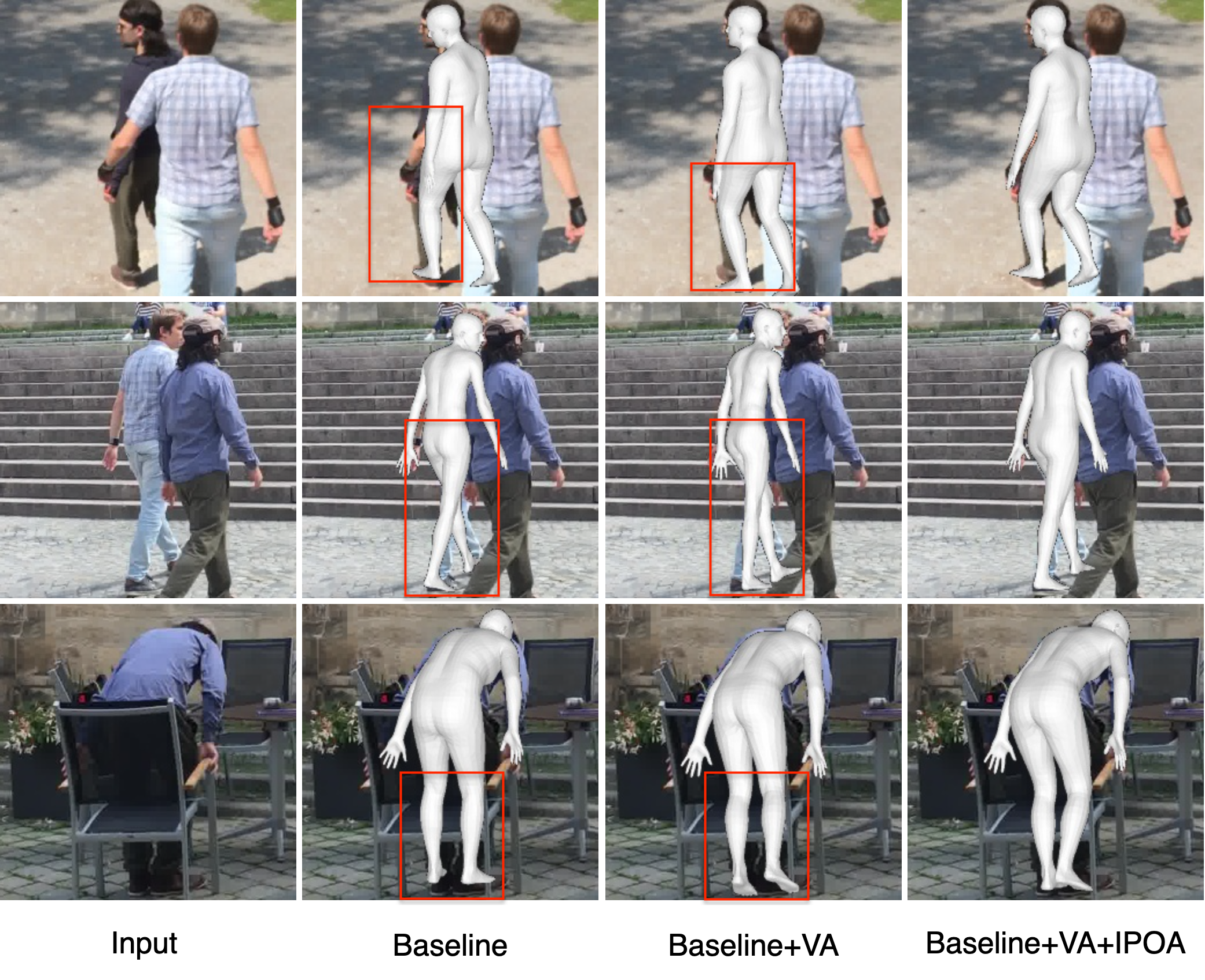}}
    \caption{Effects of view augmentation (VA) and inter-person occlusion augmentation (IPOA). The model with both VA and IPOA yields the best performance.}
    \label{fig:dif_methods}
\end{figure}

\noindent\textbf{View Augmentation and Inter-person Occlusion Augmentation.} We also investigate the effects of view augmentation (VA) and inter-person occlusion augmentation (IPOA). As shown in Table \ref{3dpw}, our method can improve the accuracy under both evaluation standards. Some qualitative results of adding the VA and IPOA modules are visualized and compared in Fig. \ref{fig:dif_methods}. The model with both VA and IPOA can handle unusual viewpoints and severe occlusions well, which yields the best performance.

{
\begin{table}[!htbp]
\caption{Ablation study on silhouette as the intermediate representation.}
\begin{tabular}{lllll}
\hline
\multirow{2}{*}{Method} & \multicolumn{2}{c}{3DPW}                                    & \multicolumn{2}{c}{SSP}                                   \\
                        & \multicolumn{1}{c}{PVE\_PA } & \multicolumn{1}{c}{MPJPE\_PA } & \multicolumn{1}{c}{PVE\_T\_SC} & \multicolumn{1}{c}{mIOU$\uparrow$} \\ \cline{2-5} 
Keypoint                & 78.8                        & 61.9                          & 18.3                           & 0.57                     \\
Keypoint+Silhouette     & \textbf{72.8}               & \textbf{57.9}                 & \textbf{14.5}                  & \textbf{0.67}            \\ \hline
\end{tabular}
\label{tab:ks}
\end{table}
\vspace{-0.2cm}
}

\begin{table}[!htbp]
\caption{Ablation study of VA and IPOA modules on 3DPW. The Baseline with * means that ResNet-50 is the encoder.}
\begin{center}

\begin{tabular}{@{}lcc@{}}
\toprule
\multirow{2}{*}{Method}         & \multicolumn{2}{c}{3DPW} \\
                                & PVE\_PA    & MPJPE\_PA   \\ \midrule
Baseline                        & 77.8       & 62.1        \\
Baseline+VA               & 75         & 59.4        \\
Baseline+VA+IPOA & 73.2       & 58.6       \\ 
Baseline*+VA+IPOA & \textbf{72.8}       & \textbf{57.9}        \\ \bottomrule
\end{tabular}
\label{3dpw}
\end{center}
\end{table}
\vspace{-0.2cm}

\begin{table}[!htbp]
\vspace{-0.2cm}
\caption{ Evaluation on the 3DPW dataset using ground-truth keypoints and silhouette.}
\begin{center}
\begin{tabular}{lcc}
\hline
Method                          & PVE\_PA       & MPJPE\_PA     \\ \hline
Baseline                        & 44.3          & 35.0          \\
Baseline+VA                & 40.8          & 32.1          \\
Baseline+VA+IPOA & \textbf{38.1} & \textbf{30.4} \\ \hline
\end{tabular}
\label{tab:groundtruth}
\end{center}
\end{table}
\vspace{-0.2cm}

\begin{table}[!htbp]
\caption{Evaluation of the 3DPW dataset. The operation of synthesizing the silhouette can generate different levels of inter-person occlusions. Percent means the probability of performing the operation.}
\begin{center}
\begin{tabular}{@{}lcc@{}}
\toprule
Percentage & PVE\_PA & MPJPE\_PA \\ \midrule
40\%       & 76.9    & 62.1      \\
80\%       & 74.8    & 60.3      \\
100\%      & \textbf{73.2}    & \textbf{58.6}      \\ \bottomrule
\end{tabular}
\label{percentage}
\end{center}
\end{table}


\begin{table}[!htbp]
\caption{Ablation study on adjusting the confidence threshold of the keypoint detection for the 3DPW dataset.}
\begin{center}
\begin{tabular}{lccccccc}
\hline
Threshold & 0    & 0.1  & 0.2  & 0.3  & 0.4  & 0.5  & 0.6  \\ \hline
PVE\_PA   & 73.9 & 73.7 & 73.5 & 73.3 & \textbf{73.2} & 73.3 & 73.9 \\ \hline
MPJPE\_PA & 59.3 & 59.1 & 58.8 & 58.7 & \textbf{58.6} & 58.7 & 59.3 \\ \hline
\end{tabular}
\label{confidence}
\end{center}
\end{table}
\vspace{-0.2cm}

\noindent\textbf{Experiments with Ground Truth 2D Inputs.}  We also report the performance of the proposed method using ground-truth keypoints and silhouettes. The results are shown in Table \ref{tab:groundtruth}. Compared with noisy 2D keypoint detection and silhouette,  the input of 2D ground-truth improves the accuracy by a large margin. This indicates that if the proposed framework is equipped with a better 2D keypoint detector and silhouette segmentation method, much better results will be obtained.

We also evaluate the effects of using different percentages of data for inter-person occlusion synthesis. 40, 80 and 100 percentage of the data are processed respectively, and the corresponding results are compared in Table \ref{percentage}. Using 100 percentage of training data to synthesize occlusions, the best accuracy is achieved on the 3DPW dataset. This reasons well since occlusions are extremely common in 3DPW.

\begin{figure}[htbp]
    \centerline{\includegraphics[width=1.0\linewidth]{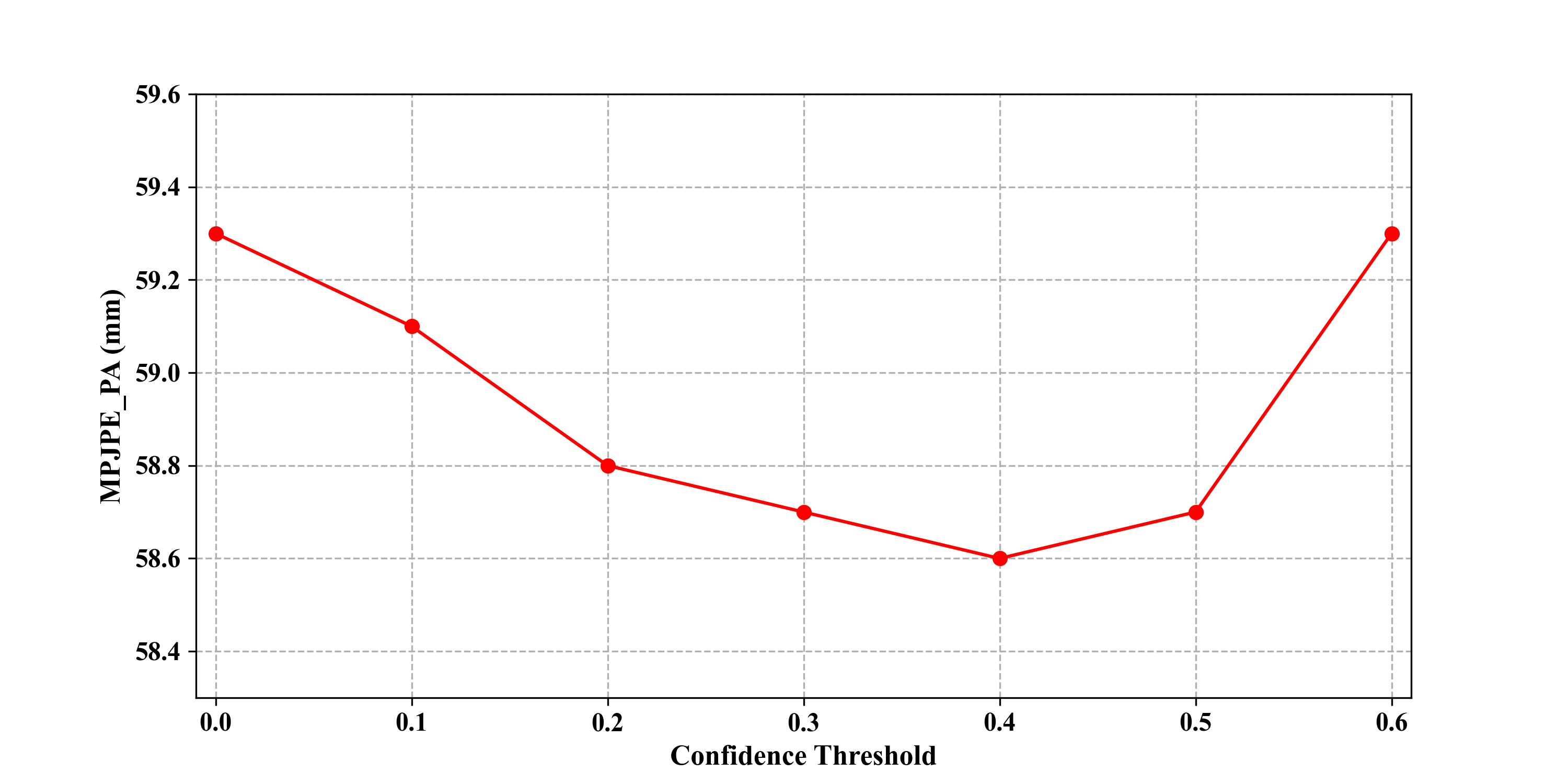}}
    \caption{MPJPE\_PA on 3DPW with different confidence thresholds.}
    \label{fig:confidence}
\end{figure}

In the 2D keypoint detector, each joint prediction is associated with a confidence score. When the predicted confidence score is above a threshold, the joint is considered detected correctly. The keypoints with confidence lower than the threshold are removed. We test the effects of adjusting the confidence threshold.  The results on 3DPW are shown in Table \ref{confidence}.  For samples with severe occlusions, by removing the poorly detected 2D keypoints properly, better results are achieved. The line chart about the MPJPE\_PA is shown in Fig. \ref{fig:confidence}. Empirically, setting the threshold to be 0.4 yields the best performance.

\section{Conclusions}
\IEEEPARstart{I}{n} this paper, we propose a new framework that synthesizes occlusion-aware silhouette and 2D keypoints representation to overcome data scarcity when regressing the human pose and shape, which proves to work well on inter-person occlusions as well as other occlusion cases. Differentiable rendering is integrated to enable silhouette supervision on the fly, contributing to higher accuracy in the pose and shape estimation. A new method for synthesizing keypoints-and-silhouette-driven data in panoramic viewpoints is presented to increase viewpoint diversity on top of existing dataset. Thanks to these strategies, the experimental results demonstrate that the proposed framework is among the start-of-the-art methods on the 3DPW and 3DPW-Crowd datasets in terms of pose accuracy. Although it is unfair to compare with these methods using paired training data, the proposed method evidently outperforms Mesh Transformer \cite{lin2021end-to-end}, 3DCrowdNet \cite{choi20213dcrowdnet} and ROMP \cite{ROMP} in terms of shape estimation. Top performance is also achieved on SSP-3D in terms of human shape estimation accuracy. The occlusion-aware synthetic data generation strategy, neural-mesh-renderer-enabled silhouette supervision and viewpoint augmentation strategy can be also applied to other approaches for 3D human pose and shape estimation to improve the overall performance of the community.

\noindent\textbf{Future Work.}  Our occlusion-aware method can be extended to the crowded people case. Video information can also be considered to obtain more consistent and accurate human pose and shape. Additionally, compared with other methods, limited training data is used for now. In the next step, more datasets like Human3.6M and COCO can be added in the training procedure to train the model.
\section*{Acknowledgment}
 The authors would like to thank the anonymous reviewers for providing constructive suggestions and comments, that contributed to highly improve this work. The authors would like to thank the authors of 3DCrowdNet, Dr. Hongsuk Choi et al., for providing the experimental results on the SSP-3D dataset in Fig. \ref{fig:ssp}.





\bibliographystyle{IEEEtran}
\bibliography{ref}
%



%

\begin{IEEEbiography}[{\includegraphics[width=1in,height=1.25in,clip,keepaspectratio]{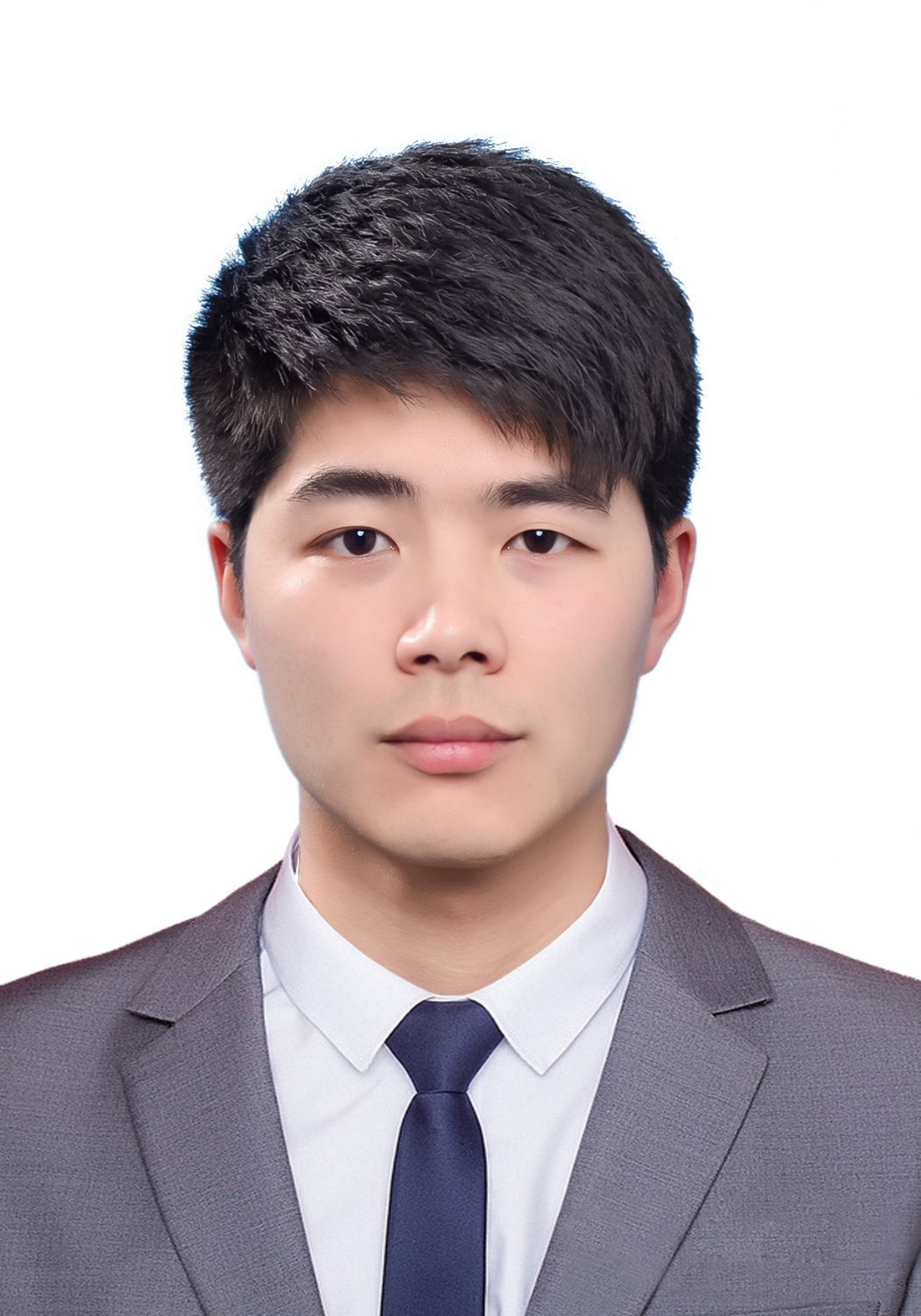}}]{Kaibing Yang} received the Bachelor's degree in Electronic Information Engineering in 2017 from Anhui Polytechnic University, P.R. China. He is currently a graduate student in the College of Computer Science and Technology, Hangzhou Dianzi University. His research interests are on geometric modeling and computer vision.
\end{IEEEbiography}

\begin{IEEEbiography}[{\includegraphics[width=1in,height=1.25in,clip,keepaspectratio]{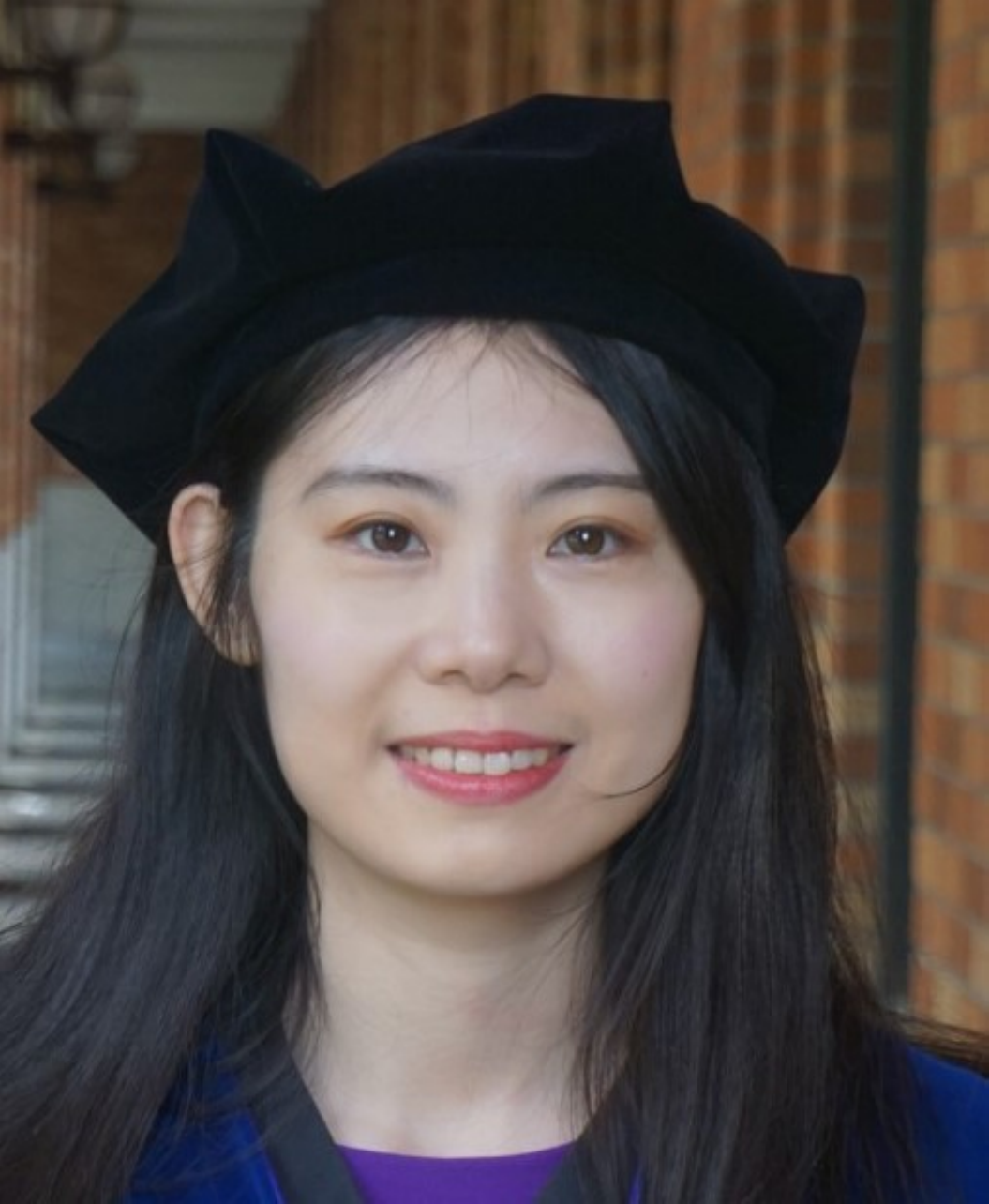}}]{Renshu Gu}
received the B.Sc. degree from Nanjing University in 2011, and the Master's and Ph.D. degrees from University of Washington in 2020, advised by Professor Jenq-neng Hwang. She is now with the College of Computer Science and Technology, Hangzhou Dianzi University, P.R. China. Her research interests include image/video analytics, computer vision and multimedia. Dr. Gu is a member of IEEE.
\end{IEEEbiography}

\begin{IEEEbiography}[{\includegraphics[width=1in,height=1.25in,clip,keepaspectratio]{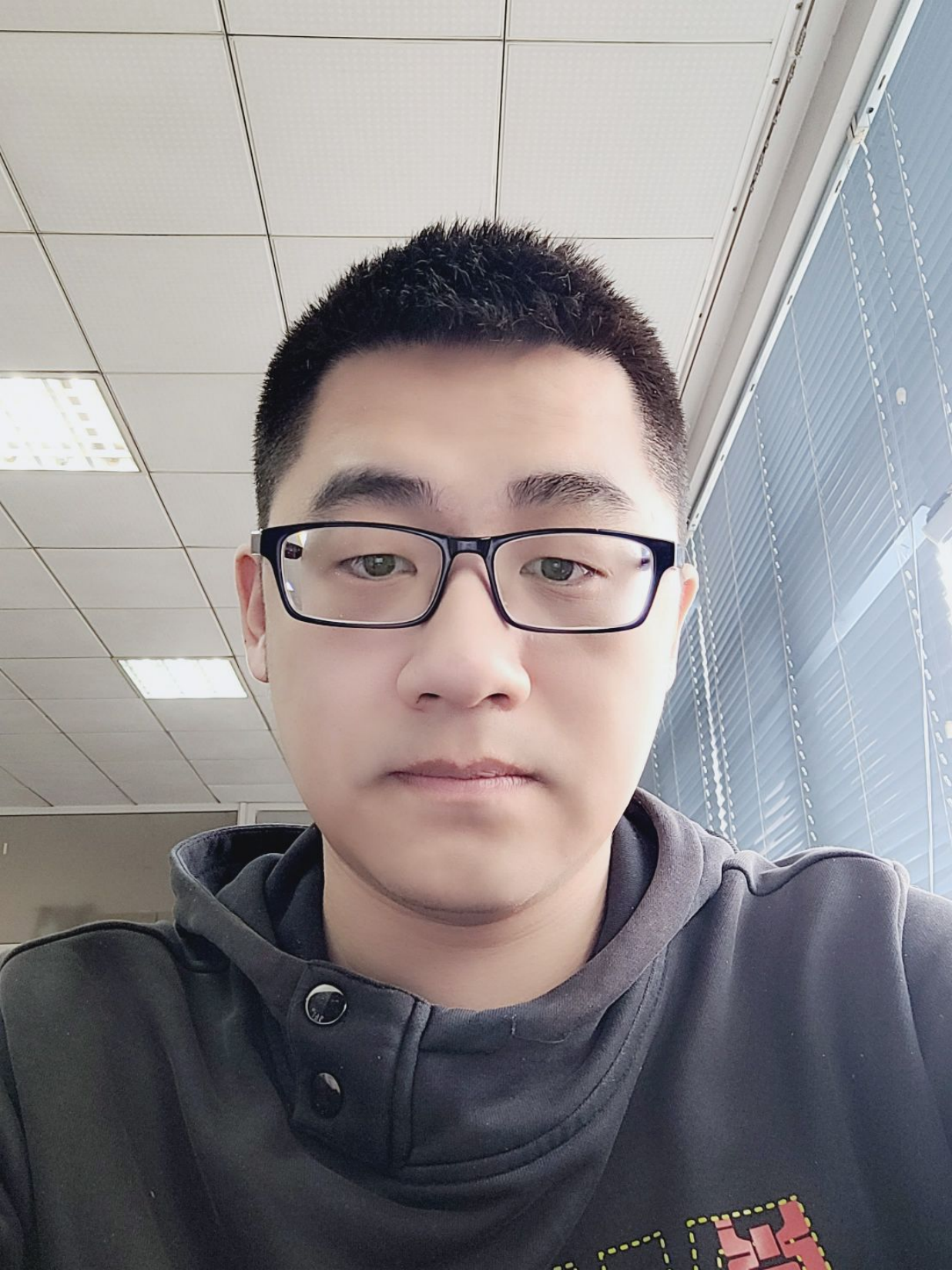}}]{Maoyu Wang} graduated from Harbin Institute of Technology in 2016 with a bachelor's degree in Internet of Things Engineering. He is currently a graduate student in Hangzhou Dianzi University ITMO Joint Institute. His research interests include geometric modeling and 3D reconstruction.
\end{IEEEbiography}

\begin{IEEEbiography}[{\includegraphics[width=1in,height=1.25in,clip,keepaspectratio]{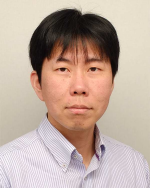}}]{Masahiro Toyoura}
received the B.Sc. degree in Engineering, M.Sc. and Ph.D. degrees in Informatics from Kyoto University in 2003, 2005 and 2008 respectively. 
He is currently an Associate Professor at Department of Computer Science and Engineering, University of Yamanashi, Japan. 
His research interests are digital fabrication, computer and human vision. He is a member of IEEE and ACM.
\end{IEEEbiography}

\begin{IEEEbiography}[{\includegraphics[width=1in,height=1.25in,clip,keepaspectratio]{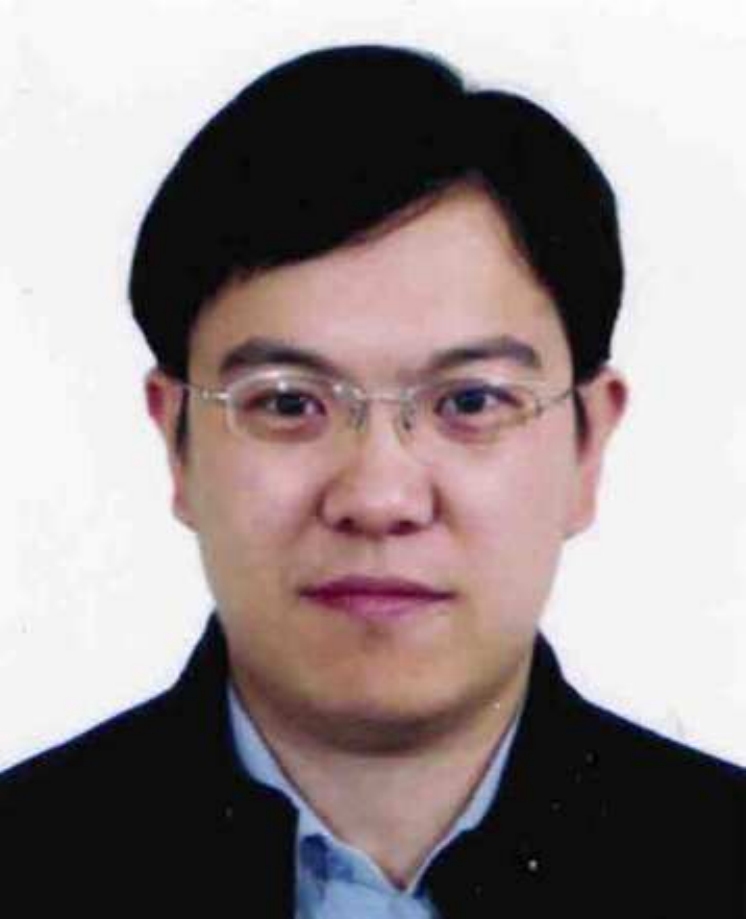}}]{Gang Xu}
received the B.Sc. degree in Mathematics from Shandong University in 2003,  Ph.D. degrees in Mathematics from Zhejiang University in 2008. 
He is currently a full Professor at College of Computer Science and Technology, Hangzhou Dianzi University, P.R. China.
His research interests are geometric modeling and simulation, computer graphics and 3D visual computing. Dr. Xu is a distinguished member of China Computer Federation. 
\end{IEEEbiography}






\end{document}